\documentclass[10pt,twocolumn,letterpaper]{article}

\usepackage{cvpr}
\usepackage{times}
\usepackage{epsfig}
\usepackage{graphicx}
\usepackage{amsmath}
\usepackage{amssymb}
\usepackage[utf8]{inputenc}

\usepackage[pagebackref=true,breaklinks=true,letterpaper=true,colorlinks,bookmarks=false]{hyperref}

\cvprfinalcopy 

\begin{document}

\title{3D Reconstruction of Incomplete Archaeological Objects Using a Generative Adversarial Network}

\author{Renato Hermoza \hspace{2cm} Ivan Sipiran \\
Pontificia Universidad Católica del Perú, Lima, Peru\\
{\tt\small renato.hermoza@pucp.edu.pe}
}

\maketitle

\begin{abstract}
   
We introduce a data-driven approach to aid the repairing and conservation of archaeological objects: ORGAN, an object reconstruction generative adversarial network (GAN). By using an encoder-decoder 3D deep neural network on a GAN architecture, and combining two loss objectives: a completion loss and an Improved Wasserstein GAN loss, we can train a network to effectively predict the missing geometry of damaged objects. As archaeological objects can greatly differ between them, the network is conditioned on a variable, which can be a culture, a region or any metadata of the object. In our results, we show that our method can recover most of the information from damaged objects, even in cases where more than half of the voxels are missing, without producing many errors.
   
\end{abstract}

\section{Introduction}

\begin{figure}[t]
\begin{center}
	\includegraphics[width=0.8\linewidth]{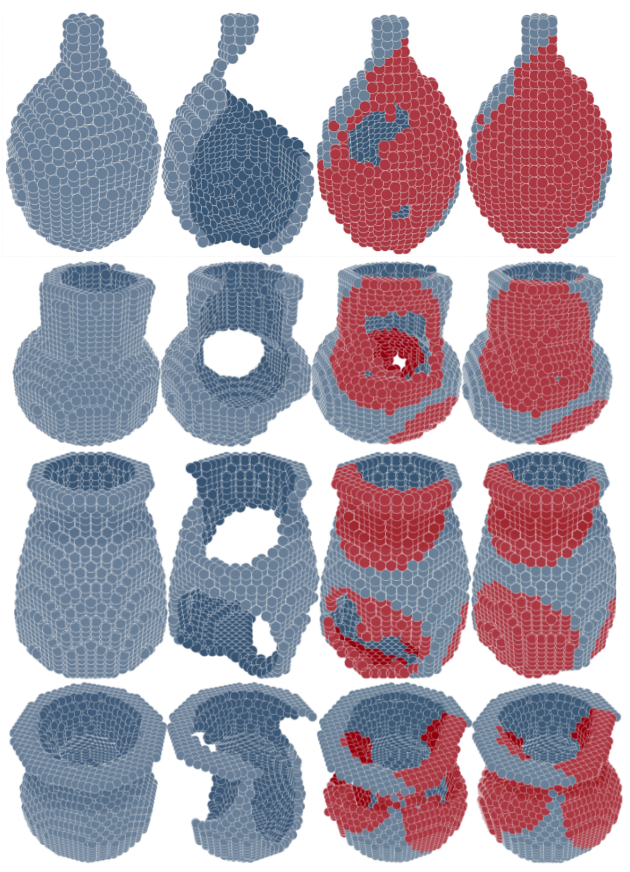}
\end{center}
   \caption{From left to right: complete objects, objects with simulated fractures, reconstruction from ORGAN and a second iteration with ORGAN.}
\label{fig:example}
\end{figure}

During archaeological excavations, it is common to find fractured or damaged objects. The process to repair and conserve these objects is tedious and delicate, objects are often fragile and the time for manipulation must be short. With the recent progress in geometry processing and shape analysis, one can address the repair problem from a computational perspective. The process starts with a 3D scanning of the object. Then, an algorithm analyzes the 3D shape to guide the conservation process. Previous experience shows that unsupervised shape analysis to repair damaged objects give good approximations to conservators, and therefore reduce the workload and time of the processing \cite{Papaioannou:2017:ROC:3068422.3009905}. 

The main problem is the prediction of missing geometry of damaged objects. Current methods assume that man-made objects exhibit some kind of structure and regularity \cite{Mitra2013}. The most common type of structure used is symmetry. If an algorithm can detect symmetries in the object, we can apply the symmetric transformation to create what is missing. Although this approach is a promising direction, there are still some drawbacks: 1) If the object is too damaged, the symmetries cannot be recovered from the object itself. 2) The computational time to search for symmetries is still high.

Deep learning techniques have proved to be highly successful in processing 3D voxelized inputs~\cite{wu_3d_2015,dai_shape_2016} and has also been recently used with generative adversarial networks (GANs)~\cite{goodfellow_generative_2014} architectures~\cite{wu_learning_2016}. We hypothesize that the aforementioned drawbacks can be addressed by a data-driven approach. It means we can learn the structure and regularity from a collection of complete known objects (in training time) and use them to complete and repair incomplete damaged objects (in testing time).

In this work we propose an object reconstruction generative adversarial network (ORGAN), for which we employ a 3D convolutional neural network (CNN) with skip-connections, as a generator on a Conditional GAN (CGAN)~\cite{mirza_conditional_2014} architecture. With two optimization targets: a mean absolute error (MAE) and an Improved Wasserstein GAN (IWGAN)~\cite{gulrajani_improved_2017} loss, the final model is encouraged to find solutions that resemble the structure of real objects. An example of a reconstructed object is shown in Figure~\ref{fig:example}. The code for the project is publicly available on a GitHub repository~\footnote{https://github.com/renato145/3D-ORGAN}.

\section{Related work}

Shape completion has gained important attention in recent years. In consequence, many approaches have been proposed so far. Pauly \etal~\cite{Pauly2005} proposed to complete a 3D scans using similar objects from a shape repository. A post-processing step of non-rigid alignment fix the transitions between the input geometry and the generated geometry. On the other hand, Huang \etal~\cite{Huang2012} computed feature-conforming fields which were used to complete missing geometry. Another interesting approach is the use of local features to guide the process of completion. For example, Harary \etal~\cite{Harary2014a} proposed to transfer geometry between two 3D objects using a similarity assessment on Heat Kernel Signatures\cite{Sun2009}. Likewise, Harary \etal~\cite{Harary2014b} proposed to complete an object with knowledge extracted from curves around the missing geometry.

An important concept that has been used to synthesize geometry is the symmetry. If one can get the information about the symmetry of an object, that information could be used to replicate portions of the object until completing it. Thrun and Wegbreit\cite{Thrun2005} proposed to complete partial scans using probabilistic measure to score the symmetry of a given object. Similarly, Xu \etal~\cite{Xu2009} designed an method to find the intrinsic reflectional symmetry axis of tubular structures, and therefore they used that information to complete human-like 3D shapes. More related to archaeological objects, Sipiran \etal~\cite{Sipiran2014} defined a strategy to find symmetric correspondences in damaged objects, which were used later to synthesize the missing geometry. Likewise, Mavridis et al.\cite{Mavridis:2015} formulated the problem of symmetry detection as an optimization problem with sparse constraints. The output of this optimization provided good hints for the process of completion of broken archaeological objects. More recently, Sipiran\cite{Sipiran_2017_ICCV} described an algorithm to determine the axial symmetry of damaged objects. This symmetry was subsequently used to restore objects with good precision.
\subsection{Deep learning methods}

With the availability of 3D shape databases \cite{chang_shapenet:_2015,koutsoudis_3d_2010,wu_3d_2014}, deep learning approaches have started to being applied on tasks involving 3D data. In Wu \etal~\cite{wu_3d_2015} a 3D convolutional neural network (CNN) is proposed for classification and shape completion from 2.5D depth maps, more recently and relevant to our case, Dai \etal~\cite{dai_shape_2016} proposes an encoder-predictor network (which follows the idea of an autoencoder) for the task of shape completion.

Recent advances in generative models with the use of generative adversarial networks (GANs) \cite{goodfellow_generative_2014} have shown an effective aid in tasks that require the recuperation of missing information while giving plausible-looking outputs. By adding a GAN loss to our model, the network is encouraged to produce outputs that reside on the manifold of the trained objects. Some examples can be seen on tasks like: super-resolution \cite{ledig_photo-realistic_2016}, image completion \cite{yeh_semantic_2016} and 3D object reconstruction from 2D images \cite{wu_learning_2016}.

One of the drawbacks of GAN models is the instability on their training, an alternative to traditional GAN is the Wasserstein GAN algorithm (WGAN) \cite{arjovsky_wasserstein_2017} which shows an improved stability by minimizing the Wasserstein distance between distributions instead of the Kullback-Leibler divergence (KL), but this algorithm forces the critic to model only K-Lipschitz functions by clamping the weights to a fixed box. A new Improved Wasserstein GAN (IWGAN) \cite{gulrajani_improved_2017} proposes an alternative method for enforcing the Lipschitz constraint, penalizing the norm of the gradient of the critic with respect to its input, resulting in faster convergence and higher quality samples.

\section{Method}

The goal of our method is to take a 3D scan of a fractured object as input and predict the complete object as output. To achieve this, we use a shape completion network that represents 3D objects as a voxel grid of size $32^3$, the completion network is then taken as the generator $G$ in a GAN architecture. The final training objective combines the completion network loss $\mathcal{L}_{comp}$ and the adversarial loss $\mathcal{L}_{adv}$.

\subsection{Data generation} \label{s:datagen}

In order to train the network, pairs of fractured and complete objects are needed, to simulate fractures from complete objects we sample $n$ random voxels from the occupied grid, then, at the each $n$ voxel a fracture of random size $m_n$ is created, having a probability $p$ of having a spherical shape, and $p-1$ of having a cubic shape. All our models were trained with $p=0.75$, $n$ sampled from 1 to 4 and $m$  from 3 to 6.

\subsection{Shape completion network}

\begin{figure*}
\begin{center}
\includegraphics[width=0.8\linewidth]{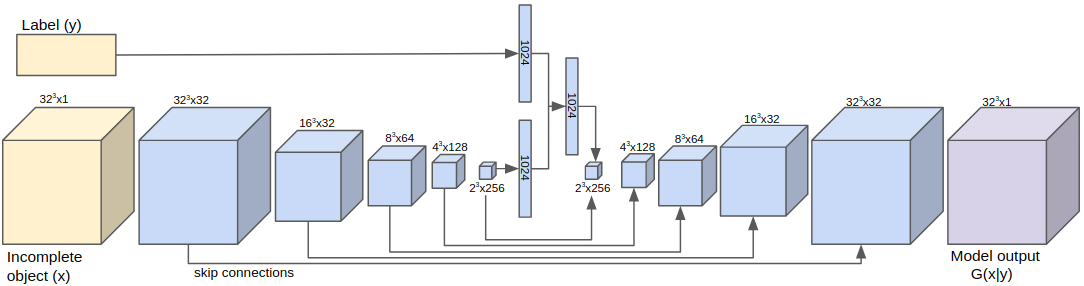}
\end{center}
   \caption{Network architecture for the generator.}
\label{fig:g}
\end{figure*}

\begin{figure}[t]
\begin{center}
	\includegraphics[width=0.8\linewidth]{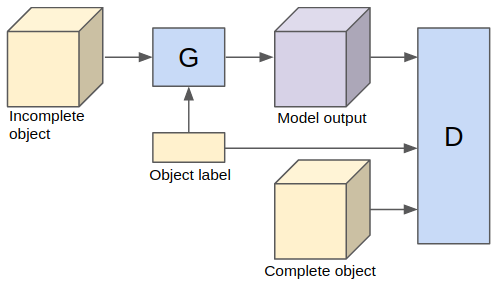}
\end{center}
   \caption{Reconstruction GAN architecture, conditioned on the object label.}
\label{fig:gan}
\end{figure}

\begin{figure}[t]
\begin{center}
	\includegraphics[width=0.8\linewidth]{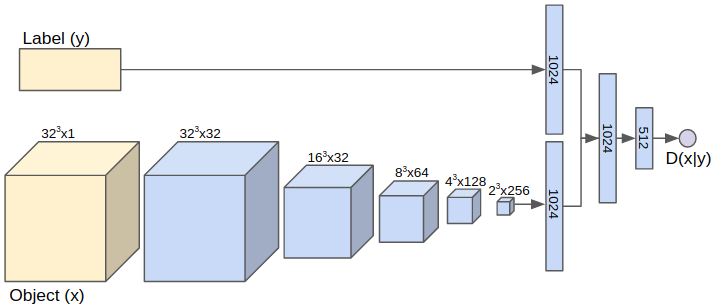}
\end{center}
   \caption{Network architecture for the discriminator.}
\label{fig:d}
\end{figure}

\begin{figure}[t]
\begin{center}
	\includegraphics[width=0.8\linewidth]{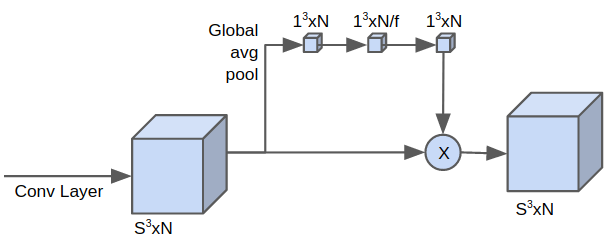}
\end{center}
   \caption{3D Squeeze-and-Excitation block.}
\label{fig:se}
\end{figure}

The network, illustrated on Figure~\ref{fig:g}, starts with a 3D encoder which compresses the input voxel grid using a series of 3D convolutional layers, the compressed hidden values are then concatenated with the embedded information about the input class label and finally a 3D decoder uses 3D transposed convolutional layers to predict the $32^3$ voxel output. Similarly to U-net architecture~\cite{ronneberger_u-net:_2015} and also used on Dai \etal~\cite{dai_shape_2016} for 3D voxels, we add skip connections on the decoder part of the network, concatenating the output of the transposed convolutions with the corresponding outputs of the encoder layers, this way we double the feature map size and allow the network to propagate local structure of the input data in the generated output.

All the layers use ReLU activations and batch normalization, with the exception of the last one that uses tanh activation and no batch normalization. After each convolutional operation a 3D Squeeze-and-Excitation (SE) block~\cite{hu_squeeze-and-excitation_2017} is applied with a reduction ratio $r=16$ as used in Hu \etal~\cite{hu_squeeze-and-excitation_2017}, see Figure~\ref{fig:se}. To train the network we use a L1-norm as the completion loss:

\begin{equation}
\mathcal{L}_{comp} = |x_t - G(x_i|y)|
\label{eq:lcomp}
\end{equation}

Where $x_t$ is the target sample, $x_i$ is the incomplete object, $y$ is the object label and $G$ is the completion network

\subsection{Adversarial network architecture}

As proposed on Goodfellow \etal~\cite{goodfellow_generative_2014}, the GAN algorithm consists of a generator $G$ and a discriminator $D$, where $G$ captures the data distribution and $D$ estimates the probability that a sample came from the training data rather than G. And following the idea of Conditional Generative Adversarial Nets (CGAN)~\cite{mirza_conditional_2014}, we extend our GAN to a conditional model, by feeding $G$ and $D$ information about class labels as an additional input layer as showed on Figures~\ref{fig:g}, \ref{fig:gan} and \ref{fig:d}.

$D$ starts similarly to $G$, compressing the voxels input and concatenating the class label, then is followed by fully connected layers, as showed on Figure~\ref{fig:d}. All layers of $D$ use Leaky ReLU activation functions and no batch normalization, with the exception of the last fully connected layer, that outputs a single value with no activation function. The 3D convolutional layers are, as in $G$, followed by a SE block.

Combining the ideas of IWGAN and CGAN, we define $D$ loss function as:

\begin{equation}
\begin{aligned}
\mathcal{L}_{adv}^D &= \mathbb{E}_{\hat{x} \sim P_g}[D(\hat{x}|y)]
                       - \mathbb{E}_{x \sim P_r}[D(x|y)] \\
               &\qquad + \lambda \mathbb{E}_{\hat{x} \sim P_x}[(||\nabla_{\hat{x}} D(\hat{x}|y)||_2 - 1)^2]
\end{aligned}
\label{eq:dgan}
\end{equation}

Where $\lambda$ is the gradient penalty coefficient, $y$ is the class label data, $P_g$ is the generator distribution, $P_r$ is the target distribution and $P_x$ is the distribution sampling uniformly on straight line between $P_g$ and $P_r$. We use $\lambda=10$ as proposed on Gulrajani \etal~\cite{gulrajani_improved_2017}.

Finally, to define $G$ loss function, we combine a typical WGAN loss with Equation~\ref{eq:lcomp}:

\begin{equation}
\mathcal{L}_{adv}^G = D(\hat{x}|y) + k\mathcal{L}_{comp}
\label{eq:ggan}
\end{equation}

Where $k$ controls the learning contribution of the completion loss. The final model uses $k=100$.

\section{Experiments}

\subsection{Data} \label{s:data}

We perform the validations of the models and hyperparameter search using the ModelNet10 dataset~\cite{wu_3d_2014}, a subset of a large 3D CAD model dataset (ModelNet), containing 10 classes (bathtub, bed, chair, desk, dresser, monitor, nightstand, sofa, table and toilet) divided on 3991 objects as the train set and 908 as the test set.

For the final model, we built a custom dataset, starting with ModelNet10 and adding an 11th class of archaeological looking objects, as the objective of this model is to reconstruct damaged archaeological objects. This new class contains 659 handpicked objects from ModelNet40 and 492 handpicked objects 3D Pottery dataset~\cite{koutsoudis_3d_2010}. The resulting dataset is divided into 4923 objects for the train set and 1127 for the test set. Finally, we show how the model performs on real fractured objects, obtained from 3D scans of archaeological objects from the Larco museum~\footnote{http://www.museolarco.org}.

For all the experiments on this work we voxelize each object at a resolution of $32^3$ voxels using Binvox~\cite{min_binvox_2004,nooruddin_simplification_2003} and scale the binary voxels to $[-1,1]$.

\subsection{Training details}

The models were implemented using Keras framework~\cite{chollet_keras_2015} with Tensorflow~\cite{martin_abadi_tensorflow:_2015} as backend. We trained all the experiments on a NVIDIA TITAN Xp using a batch size of 64, with the generator model training every 5 batches and the discriminator model training every batch. For optimization we use Adam~\cite{kingma_adam:_2015} with $\alpha=0.0001$, $\beta_1=0.5$ and $\beta_2=0.9$, as proposed on Gulrajani \etal~\cite{gulrajani_improved_2017}. The final model is trained for 400 epochs.

\subsection{Performance of the final network}

\begin{table}
\begin{center}
\caption{L1 loss results on different network settings against the test data set. Each setting was run for 100 epochs.}
\begin{tabular}{c|c|l}
\hline
Skip-connections & Squeeze-and-excite & L1 loss \\
\hline
No & No & 0.0611 \\
Yes & No & 0.0061 \\
Yes & Yes & 0.0057 \\
\hline
\end{tabular}
\label{t:tests}
\end{center}
\end{table}

\begin{table}
\begin{center}
\caption{L1 loss results by class label for the test data set.}
\begin{tabular}{l|l|l}
\hline
Label & Input loss & Output loss \\
\hline
Archeology & 0.0209 & 0.0077 \\
Bathtub & 0.0180 & 0.0068 \\
Bed & 0.0235 & 0.0058 \\
Chair & 0.0185 & 0.0103 \\
Desk & 0.0202 & 0.0114 \\
Dresser & 0.0192 & 0.0027 \\
Monitor & 0.0181 & 0.0062 \\
Nightstand & 0.0203 & 0.0052 \\
Sofa & 0.0231 & 0.0063 \\
Table & 0.0129 & 0.0039 \\
Toilet & 0.0216 & 0.0100 \\
\hline
\end{tabular}
\label{t:results1}
\end{center}
\end{table}

\begin{figure}
\begin{center}
\includegraphics[width=0.8\linewidth]{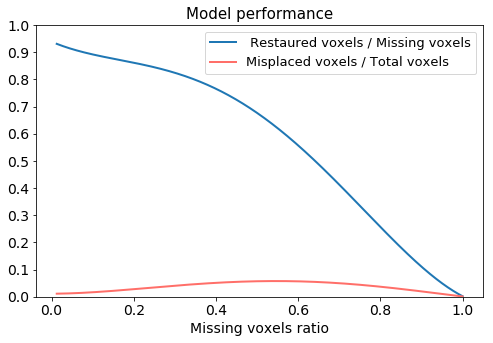}
\end{center}
   \caption{Model average performance against different randomly generated fractures. The number and size of the fractures vary between 1 and 15. We can see that even when 40\% of the voxels are missing, we can still recover 80\% of the information.}
\label{fig:psizes}
\end{figure}

\begin{figure*}
\begin{center}
\includegraphics[width=0.8\linewidth]{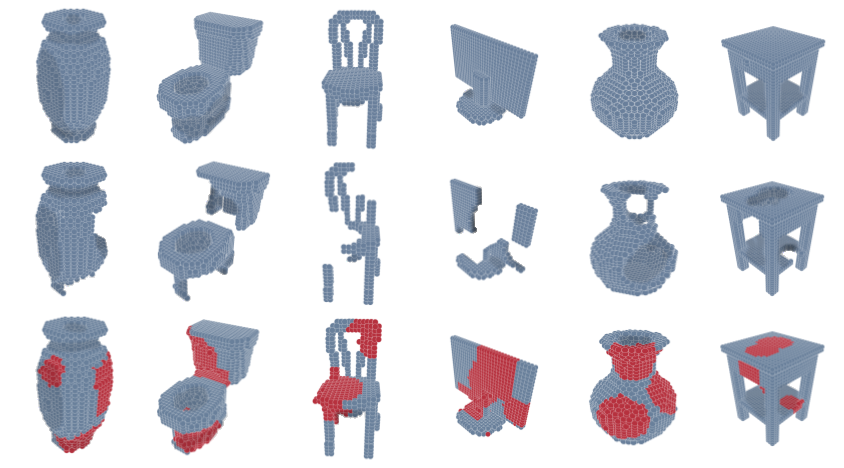}
\end{center}
   \caption{Results on different labels with a maximum fracture size of 6. First row: Complete objects. Second row: Objects with fractures. Third row: Results from our model.}
\label{fig:res1}
\end{figure*}

\begin{figure*}
\begin{center}
\includegraphics[width=0.75\linewidth]{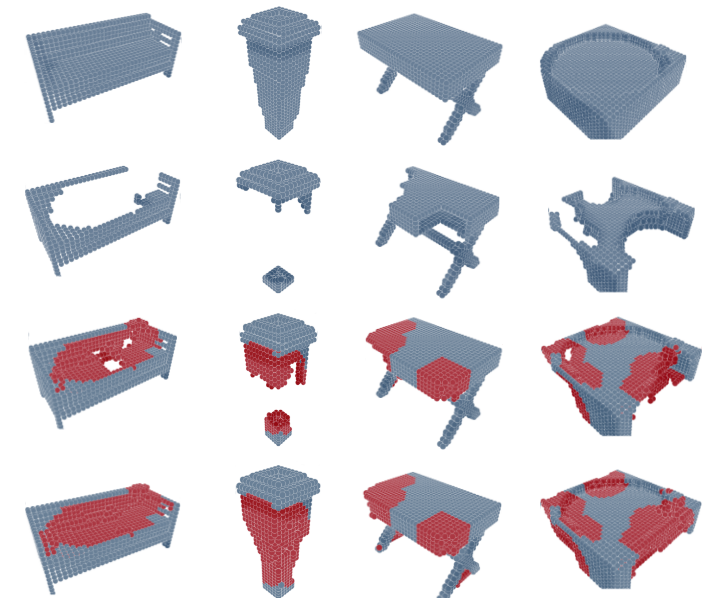}
\end{center}
  \caption{Results on different labels with a maximum fracture size of 12. First row: Complete objects. Second row: Objects with fractures. Third row: Results from our model. Fourth row: A second iteration with our model.}
\label{fig:res12}
\end{figure*}

\begin{figure}
\begin{center}
\includegraphics[width=0.8\linewidth]{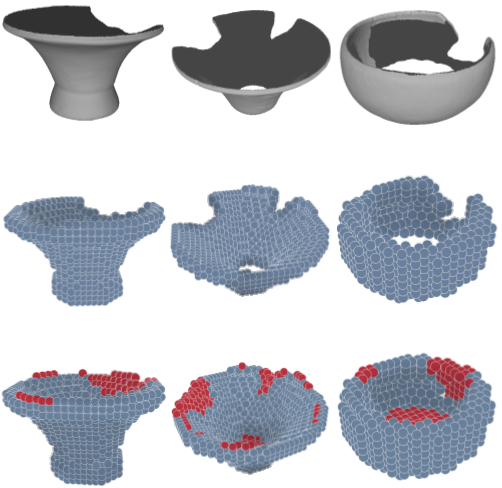}
\end{center}
   \caption{Results from scanned archaeological objects. First row: Scanned archaeological objects. Second row: Obtained voxels. Third row: Results from our model.}
\label{fig:res21}
\end{figure}

\begin{figure}
\begin{center}
\includegraphics[width=0.7\linewidth]{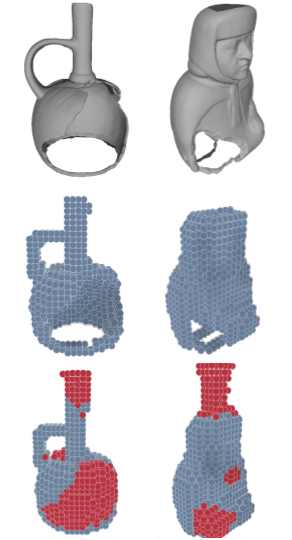}
\end{center}
   \caption{Unexpected artifacts being reconstructed. First row: Scanned archaeological objects. Second row: Obtained voxels. Third row: Results from our model.}
\label{fig:res22}
\end{figure}

In order to choose the final network configuration, we trained different settings for 200 epochs each. The results show that using skip-connections is important for the network performance, and SE blocks can also give a performance boost, as seen in Table~\ref{t:tests}.

When using the model at inference time, only the generator is used. We tested the trained model on our custom dataset, reducing significantly the loss as seen on Table~\ref{t:results1}. On Figure~\ref{fig:psizes} we show the performance against different fragment sizes, it can be seen that the model can recover most of the information, even when more than half of the voxels are missing, without producing many misplaced voxels. Results on different labels from the test data set are shown on Figure~\ref{fig:res1}. In some cases, where the number of missing voxels was greater than 2000, an additional run on the model was performed against the first result as shown on Figures~\ref{fig:example} and \ref{fig:res12}.

We also performed tests on real fractured archaeological objects, results can be seen on Figure~\ref{fig:res21}. Some of the real objects greatly differ from the ones used in trained, on Figure~\ref{fig:res22}, we show some examples of unexpected fragments being reconstructed. This happened in objects whose structure greatly differed from that of the objects used in training.

\section{Conclusion}

This paper presents a method to predict the missing geometry of damaged objects: ORGAN, an object reconstruction generative adversarial network. Our results show that we can accurately recover an object structure, even in cases where the missing information represents more than half of the input occupied voxels. When tested on real archaeological objects, we showed some cases of unexpected artifacts being reconstructed, this was expected since the objects in this cases had different structures from the ones on the training set. As the objective of this work is to aid the conservation of archaeological objects, and knowing that this objects can greatly differ from one culture to another, we prepared our method to acts as a conditional model on a variable, which can be a culture, a region or any metadata of the object. Then, an important task is to increase the amount of data we have about archaeological objects, which we leave as our future work.


{\small
\bibliographystyle{ieee}
\bibliography{Zotero1}
}












\end{document}